\documentclass{article}

\usepackage{arxiv}

\usepackage[utf8]{inputenc} 
\usepackage[T1]{fontenc}    
\usepackage{hyperref}       
\usepackage{url}            
\usepackage{booktabs}       
\usepackage{amsfonts}       
\usepackage{nicefrac}       
\usepackage{microtype}      
\usepackage{lipsum}
\usepackage{graphicx}
 \usepackage{amsmath}
 \usepackage{xcolor}
\graphicspath{ {./images/} }

\title{Graph Query Networks for Object Detection with Automotive Radar \thanks{This is the author-accepted manuscript of a paper that will appear in main proceedings of IEEE/CVF Winter Conference on Applications of Computer Vision (WACV) 2026. }}

\author{
Loveneet Saini$^{1}$ \quad Hasan Tercan$^{1}$ \quad Tobias Meisen$^{1}$\\
$^{1}$ University of Wuppertal, Germany\\
{\tt\small sainiloveneet.ls@gmail.com, \{tercan, meisen\}@uni-wuppertal.de}
}

\begin{document}
\maketitle

\begin{abstract}
Object detection with 3D radar is essential for  360$^{\circ}$ automotive perception, but radar's long wavelengths produce sparse and irregular reflections that challenge traditional grid and sequence-based convolutional and transformer detectors. This paper introduces Graph Query Networks (GQN), an attention-based framework that models objects sensed by radar as graphs, to extract individualized relational and contextual features. GQN employs a novel concept of graph queries to dynamically attend over the bird's-eye view (BEV) space, constructing object-specific graphs processed by two novel modules: EdgeFocus for relational reasoning and DeepContext Pooling for contextual aggregation. On the NuScenes dataset, GQN improves relative mAP by up to +53\%, including a +8.2\% gain over the strongest prior radar method, while reducing peak graph construction overhead by 80\% with moderate FLOPs cost. 
\end{abstract}

\section{Introduction}
\label{sec:intro}
Depth sensing sensors are central to automotive perception. While lidar has driven substantial advances, its high cost hinders large-scale deployment \cite{lidar_survey_2023}. Radar, by contrast, offers a low-cost, all-weather alternative with Doppler velocity measurement capabilities \cite{radar_classification_survey_2020}. Recent developments in 4D radars provide lidar-like performance but at a higher cost \cite{4d_radar_market_report_2024}. As a result, 3D radar, which provides range-azimuth-Doppler (RAD) measurements, has become standard in modern vehicles \cite{3d_radar_remote_sensing_2019}, offering a practical tradeoff between affordability and sensing capacity.

\begin{figure}[htbp]
	\centerline{\includegraphics[scale=0.7]{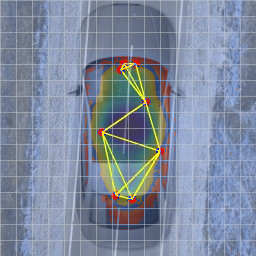}}
	\caption{Visualization of a graph query, with dynamic learning of nodes (red) and edges (yellow) guided by the attention map.}
	\label{gq}
\end{figure}

At a fundamental level, radar operates by emitting and capturing millimeter-wave reflections from object surfaces \cite{patole2017automotive}, which are processed into the RAD cube using Fast Fourier Transforms (FFTs) \cite{major2020survey}. These reflections are inherently sparse and irregular, as they depend on surface reflectivity, geometry, and orientation \cite{hasch2012millimeter}. This leads to non-deterministic and often incomplete object structures. Conventional models such as convolutional neural networks (CNNs) and Transformers impose regular grids or sequence structures, which can distort such irregularities and obscure relational and contextual cues critical for interpreting sparse object representations. To address this, we propose a graph-based approach that preserves and exploits this structural irregularity by representing each object as a graph, where nodes model object parts and edges encode spatial relationships. This allows us to extract relational features within objects and contextual features from their surroundings via attention-based reasoning. We call this graph reasoning, which is a core capability of our approach.

To effectively reason over such irregular radar data, attention-based architectures offer a natural foundation via flexible, query-driven processing. Past approaches have used 1D queries to represent object proposals \cite{carion2020end} or 2D Bird's-eye view (BEV)-aligned queries for aggregating multi-view image features \cite{li2022bevformer}. Extending this idea, we introduce graph queries (Figure \ref{gq}), wherein each query constructs an object-specific graph by attending over BEV feature space. This structured formulation is especially beneficial for radar, where object returns are sparse and fragmented. By capturing meaningful relationships and interactions, graph queries could support semantic completion from partial observations.
 
When applying graph processing over radar data, a key challenge is the computational overhead of graph construction. Full-graph construction over the input is computationally expensive. In contrast, our graph queries are initialized as empty structures that selectively populate nodes and edges via attention over the radar BEV space.  The relatively small size of each query graph compared to graphs over the entire input, as will be shown, enables efficient feature extraction using graph processing.

In this paper, we propose Graph Query Networks (GQN), a novel attention-based framework that dynamically constructs and processes graph queries for radar object detection. Each query samples nodes and constructs edges. Then, it is processed by two new components: The first is EdgeFocus, a query update operator that extracts fine-grained relational features from local edge structures. The second is DeepContext Pooling, which models object-to-object interactions by sharing information across queries. GQN acts as a plug-and-play module that can be integrated into existing detectors to enhance performance through graph reasoning. 

Finally, we extend GQN into a Unified Reasoning Architecture (URA) that integrates temporal, spatial, and graph-based feature extraction for robust radar perception. We benchmark our model on the nuScenes dataset \cite{caesar2020nuscenes} using standard protocols, and additionally evaluate GQN on a more complex proprietary radar dataset \cite{braun2021quantification}, highlighting its advantages across object categories with varying sizes and motion characteristics. In summary, the main contributions of this paper are:

\begin{itemize}
\item We propose Graph Query Networks (GQN), introducing dynamic graph queries as a flexible, low-overhead mechanism for object-centric graph reasoning in radar perception. 

\item We introduce EdgeFocus, a query update operator that captures fine-grained relational features by attending to edge interactions.

\item We present DeepContext Pooling, a context-sharing module designed to capture interactions between objects.

\item We design a Unified Reasoning Architecture by combining existing temporal and spatial modules with GQN, for robust and comprehensive radar perception.

\item We benchmark our approach on the NuScenes dataset, showing that it complements existing radar detectors and improves object detection performance.
\end{itemize}

\section{Related Works}
\label{sec:related}

\begin{figure*}[htbp]
    \centering
    \def\svgwidth{\linewidth}
    \fontsize{8}{10}\selectfont
    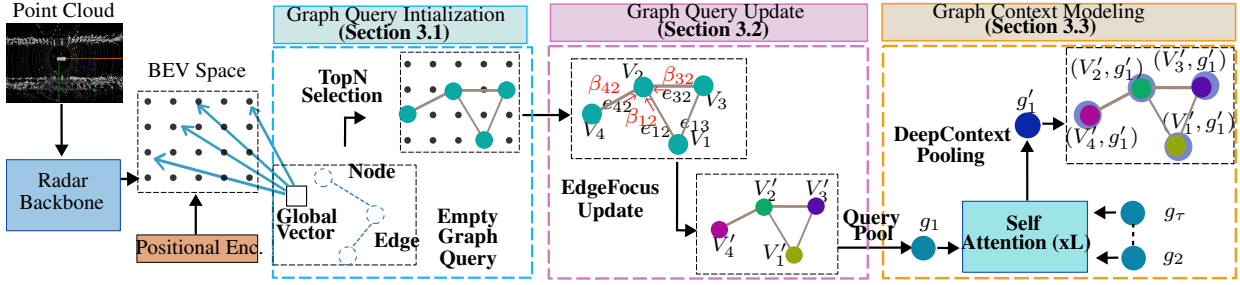
    \caption{GQN pipeline: Graph queries are instantiated over BEV features using attention guided sampling, updated via attention over edges (EdgeFocus), and enriched through context pooling and self-attention (DeepContext Pooling). \textbf{Symbols:} $V_i$: node,  $e_{ij}$: edge, $\beta_{ij}$: attention weight for edge $e_{ij}$, $g_i$: pooled graph query vector, $V'_i$: updated node, $g'_i$: updated pool vector.
 }
    \label{gqn}
\end{figure*}

\textbf{Graph-Based Methods in Camera and Lidar Domains}: 
Graph-based methods have gained traction across lidar and vision tasks for their ability to model spatial structure and part-to-part relationships. Approaches such as DGCNN \cite{wang2019dynamic}, PointGNN \cite{shi2020point}, SAT-GCN \cite{wang2023sat}, VP-Net \cite{song2023vp}, and Graph R-CNN \cite{yang2018graph} introduce dynamic or semantic graph construction over dense point clouds. Related k-NN–based frameworks, including GraphAlign and GraphAlign++ \cite{song2023graphalign,song2023graphalign++} and FocalAD \cite{sun2025focalad}, extend this principle to multimodal alignment, while methods like Graph-BEV \cite{song2024graphbev} and VisionGNN \cite{han2022vision} adapt it to camera feature space. Despite strong representational capability, graph construction can dominate runtime; \cite{wu2024point} reports up to 28\% of inference time, motivating serialized neighborhoods at the cost of adaptability.

These models typically embed graph construction within tightly coupled, end-to-end pipelines, which is an effective design for dense modalities like camera or lidar, where signal continuity and feature richness support global graph reasoning. However, directly adapting such designs to radar is non-trivial, as radar operates in a fundamentally different feature space (range, azimuth, Doppler) and exhibits extreme sparsity and irregularity. Its FFT-based processing chain and sensitivity to object orientation \cite{hasch2012millimeter}, impose unique constraints that require radar-specific design. In contrast, our approach introduces a modular, query-driven graph formulation that enables lightweight, object-specific reasoning tailored to radar's structural characteristics.

\textbf{Grid and Sequence Based Methods}: Despite radar's unique challenges, many recent methods adopt CNN or transformer-based paradigms originally designed for vision. CNN-based approaches convert radar data into regular grids and apply spatial processing techniques such as feature pyramid networks (FPN) \cite{lin2017feature}, adapting these methods to radar inputs \cite{lang2019pointpillars,musiat2024radarpillars,lee2020deep}. Transformer-based models treat BEV radar representations as token sequences with positional encodings, enabling global attention-based processing \cite{Saini_2024_CVPR,bai2021radar,peng2024transloc4d,jiang2023mlrt}. Several models integrate temporal modeling to capture motion cues \cite{li2022exploiting, yataka2024radar, yataka2024sira,saini2025attentivegru}.

While these architectures advance radar detection, they rely on location-tied features that do not explicitly model structural relationships within or across objects. As a result, these models may struggle in scenes with sparse or ambiguous measurements, where structural and contextual cues could support reliable detection. Without dedicated mechanisms to reason over such irregularities and enforce geometric consistency, these models risk overlooking critical relational structure within and across objects.

\textbf{Graph Based Methods in Radar Perception}: Graph neural networks are well-suited for radar's irregular and sparse data. Prior works 
 construct static $K$-nearest neighbor graphs (typically $K$=20)  over the entire radar point cloud to support message passing across returns \cite{fent2023radargnn,svenningsson2021radar}. However, these static full-graph designs are computationally intensive and poorly suited to dynamic object layouts. Other strategies include manually defined region-limited dynamic graphs centered on high-reflection points for classification \cite{saini2023graph}, or static graphs over RAD tensors with Region Proposal networks \cite{meyer2021graph}.

In contrast, our method introduces dynamic graph queries that selectively sample and process object-specific subgraphs directly in radar's BEV feature space. These query-specific graphs are constructed per object, enabling fine-grained reasoning with significantly lower compute cost and improved adaptability to sparse inputs. This formulation reframes graph processing as a lightweight, modular alternative to end-to-end graph pipelines, enabling integration into modern radar detectors without incurring substantial architectural overhead.

\section{Method Overview}
GQN operates atop standard radar backbones \cite{K_hler_2023,kohler2023improved,Saini_2024_CVPR,tan20223d,ulrich2022improved} and is flexibly integrated after projecting radar point clouds into the BEV space. The framework is modular and can be integrated with different BEV-based detection backbones without altering their overall architecture.

Following the formalism established in relational inductive bias literature \cite{battaglia2018relational}, we define a graph $G$  as a tuple ($u$, $V$, $E$), where:

\begin{itemize}
\item $u\in \mathbb{R}^d$ is a global feature vector representing properties shared across the graph, with $d$ as the feature dimensionality,
\item $V =\{ v_{1},\ldots,v_{N}\}$ is a set of  nodes representing entities, with cardinality  $|V| = N$,  
\item $ E = \{ (v_{1}, v_{2}),\ldots,(v_{i},v_{j})\} \subseteq V \times V $ is as set of undirected edges between node $v_{i}$ and $v_{j}$,  defined by a connectivity rule.
\end{itemize}

Building on this foundation, we define a single \textbf{Graph Query} \textbf{$G_q$} over an input space as a learnable embedding structure composed of a single global vector $u$, a set of $N$ sampled nodes, and a set of learned edges among those nodes. As illustrated on the left side of Figure \ref{gqn},  each graph query is initialized as an empty structure with a learnable global vector and placeholders for nodes and edges. These are progressively populated by sampling points from the input BEV feature space. Rather than relying on predefined neighborhoods, node selection is conditioned on the global vector, enabling each graph to adapt dynamically to the scene. This facilitate selective grouping of semantically related points while suppressing clutter and noise.  

A central challenge in radar perception lies in ensuring that queries attend to true object regions rather than background noise and clutter. To address this, GQN employs a center-based keypoint detection strategy that supervises query learning with object-level guidance. Building on this supervision, the framework unfolds as shown in Figure~\ref{gqn}: graph query initialization selects informative nodes from sparse BEV features; the EdgeFocus operator refines these nodes by modeling fine-grained relational dependencies; and DeepContext pooling exchanges evidence across queries to resolve ambiguities. Section~3.4 further extends these core stages with a multi-set design, which spans different sampling ratios to capture structure at multiple sparsity levels, and with query supervision, where global vectors are reused by the detection head to provide direct object-level learning signals.

\subsection{Graph Query Initialization}

To initialize a graph query $G_q$, we employ an attention mechanism that dynamically selects node locations in the BEV input space, conditioned on a learnable global query vector. This mechanism computes a compatibility score between the global vector and each BEV feature location, guiding the model to focus on informative regions. The entire process is fully differentiable, enabling end-to-end learning of node selection aligned with task objectives.

Following common practices in radar perception  \cite{K_hler_2023,kohler2023improved,Saini_2024_CVPR,tan20223d,ulrich2022improved}, we adopt a Pillar FeatureNet based on the PointPillars framework \cite{lang2019pointpillars} as our generic backbone. Radar reflections are first processed using a 2D Fast Fourier Transform (FFT), then projected into a BEV grid via Pillar FeatureNet to produce a BEV feature map for downstream processing.

Formally, let 
$\chi = \{x_{1}, \ldots, x_{M_{BEV}}\}, \; x_{k} \in \mathbb{R}^d$ 
denote the set of BEV state feature vectors of dimensionality $d$, with total count $M_{BEV}$. 
These are concatenated with fixed sinusoidal positional encodings \cite{vaswani2017attention}
$P = \{p_{1}, \ldots, p_{M_{BEV}}\}, \; p_{k} \in \mathbb{R}^d$, 
to form the input set.

For each graph query $G_q$ among a total of $\boldsymbol{\tau}$ queries, we compute attention scores $\alpha = \{\alpha_k\}_{k=1}^{M_{BEV}}$ between the global vector $u  \in \mathbb{R}^d$ of $G_q$ and each BEV feature $x_{k}$ as: 

\begin{equation}
\alpha_k = \frac{\exp( u^\top x_k)}{\sum_{l=1}^{M_{BEV}} \exp( u^\top x_l)}
\end{equation}

Importantly, attention is computed over the state features $x_k$, which encode semantic information from the scene, rather than the fixed positional encodings $p_k$. Using position encodings alone would bias node selection toward spatial proximity, limiting generalization and robustness to scene variations. 

Let $\mathcal{I}_{\text{top}} \subset \{1,\ldots,M_{BEV}\}$ where $|\mathcal{I}_{\text{top}}| = N $, denote the indices of the top-$N$ attention scores among $\alpha = \{\alpha_k\}_{k=1}^{M_{BEV}}$. The corresponding state-position feature pairs form the graph node set:

\begin{equation}
V = \left\{ \left( x_k, p_k \right) \,\middle|\, k \in \mathcal{I}_{\text{top}} \right\}
\end{equation}

Each selected node inherits its state feature and positional encoding from the BEV map and is assigned to a node placeholder in $G_q$. Due to permutation-invariant nature of graphs, no strict ordering is enforced and any selected node can be mapped to any placeholder. 

Edges are dynamically constructed by connecting each node to its $K$ nearest neighbors in the state feature space, following standard graph learning practice \cite{wang2019dynamic}. This yields a fully differentiable, data-adaptive graph structure that encodes both semantic saliency and geometric relationships within the scene.

\subsection{Graph Query Update}

After initializing a graph query, we process its structure using a attention-based message-passing operator to extract relational features among sampled nodes.  For each node $v_{i}$ , we define a local neighborhood $\mathcal{N}(i)$ and compute edge features for each neighbor 
$v_j \in \mathcal{N}(i)$ as: 

\begin{equation}
e_{ij} = \phi(p_{j} - p_{i}\, \| \, x_{j})
\label{edge}
\end{equation}

Here, $\phi(.)$ is a shared multi-layer perceptron (MLP), $p_i$ and $p_j$ are abstract positional encodings of the nodes, and $x_j$ is the state feature of node  $v_{j}$. By concatenating, the relative position $p_{j} - p_{i}$ with the semantic feature $x_{j}$, each edge feature encodes both geometric and feature-based cues.

Although node selection during query initialization already emphasizes informative spatial regions, the subsequent challenge is modeling the strength of relationships between nodes. For example, two nodes may be distant in position yet similar in velocity, suggesting likely membership of the same object. Since our edge feature formulation in Equation~\ref{edge} incorporates both geometric and semantic aspects, we apply attention directly over edge features to assess relational strength, shifting the focus from who the neighbor is to how the neighbor relates. While attention on edges has been explored in heterogeneous graph learning \cite{velivckovic2017graph}, our formulation, termed \emph{EdgeFocus}, adapts this principle as the query update operator within a query-centric radar pipeline, where structural meaning often lies in sparse, fragmented, and noisy returns. By operating entirely over edge features $e_{ij}$, EdgeFocus enables expressive object-aware reasoning between selected nodes. This design naturally aligns with the query-based paradigm and supports efficient query update.

To prioritize informative relationships, we compute an attention weight $\beta_{ij}$ for each edge $e_{ij}$ as:

\begin{equation}
    \beta_{ij} = \frac{\exp (q(e_{ij})^\top k(e_{ij'})) }
    {\sum\limits_{j' \in \mathcal{N}(i)}\exp(q(e_{ij})^\top k(e_{ij'}))}
\end{equation}
    
where  $q(e_{ij})$ and $k(e_{ij'})$ are learnable projections of edge features via separate MLPs. This formulation allows each edge to attend over its peers in the neighborhood.

Finally, as illustrated for node $V_2$ in the center of Figure \ref{gqn}, the updated node representation $v'_i$ is obtained by aggregating attention-weighted edge features and combining them with the original node feature $x_{i}$ using an MLP $\rho(.)$ as:

\begin{equation}
    v'_i =\rho(\sum \limits_{j \in \mathcal{N}(i)}\beta_{ij}e_{ij},x_i)
    \label{up}
\end{equation}

This results in refined node embeddings that encode structured, attention-weighted relationships among selected graph members.

\begin{figure*}[htbp]
    \centering
    \def\svgwidth{\linewidth}
    \fontsize{8}{10}\selectfont
    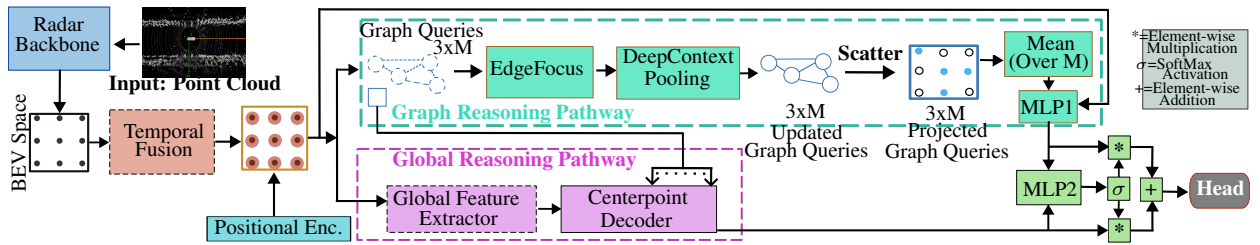
    \caption{Unified Reasoning Architecture. BEV features from a backbone and flexible temporal fusion (e.g., ConvGRU or AttentiveGRU) are processed in parallel by a GQN and a global feature extractor (e.g., Transformer or FPN) with a CenterPoint decoder. The global vectors of the graph queries are reused as decoder queries. Outputs are fused via attention and passed to a centerpoint detection head.}
    \label{full}
\end{figure*}

\subsection{Graph Context Modeling}

While the EdgeFocus operator models intra-graph structure, real-world radar scenes often require reasoning across multiple object queries, for instance, to merge parts of large objects or resolve inter-object ambiguities. To this end,  we propose a lightweight inter-query context module, DeepContext Pooling, which enables information sharing among graph queries. 

Each graph query $G_q$ is first summarized into a compact representation using adaptive max pooling over its updated node features (Equation \ref{up}):

\begin{equation}
g_q = \text{AdaptiveMaxPool}(\{v'_i\}), \quad v'_i \in G_q
\end{equation}

The resulting vector $g_q$ serves as a graph-level summary of the object or its facets encoded by query $G_q$. To model inter-object interactions, we apply iterative self-attention for $L$ steps across the set $\{g_q\}_{q=1}^\tau$ (shown on the right side of Figure \ref{gqn}), enabling information exchange among all $\tau$ graph queries:

\begin{equation}
g'_q = \text{IterativeSelfAttn}(g_q, \{g_q\}_{q=1}^\tau)
\end{equation}

This produces an updated summary $g'_q$ that incorporates contextual cues form surrounding objects or different parts of the same object identified by other queries. 

The context-enhanced vector $g'_q$ is concatenated with each node feature $v'_i \in G_q$ and passed through a shared MLP $\eta(\cdot)$ to produce context-aware node embeddings:

\begin{equation}
v''_i = \eta([v_i' \, \| \, g'_q])
\end{equation}

This query-centric formulation complements local graph processing by enabling queries to share contextual evidence, thereby aligning local features with inter-object context. This promotes structural consistency across queries and supports robust object detection.

\subsection{Graph Query Network: Multi-Set Graph Reasoning Module}

The Graph Query Network (GQN) is a modular reasoning component designed to capture relational and contextual structure in radar data. Operating on BEV features, GQN uses learnable graph queries to sample and process object-centric subgraphs, enabling targeted reasoning under sparse and noisy conditions. 

\textbf{Multi-Set Query Design:} To model relational patterns at varying sparsity, GQN adopts a multi-set query structure. Each set contains $M$ learnable graph queries, initialized with distinct sampling ratios (defined as $N \div M_{BEV}$) to represent different graph densities. For example, a three-set configuration ($\tau = 3\times M)$,  may sample 10\%, 20\%, and 30\% of BEV locations. As motivated in ablations (Section 5.4), this design enables each query group to focus on a different relational granularity. Coarse graphs focus on capturing global structure with minimal noise, while denser graphs facilitate finer reasoning over small or structurally complex objects. 

Each graph query $\{G_q\}_{q=1}^{\tau}$, samples candidate BEV nodes via attention, using its set-specific sampling ratio, and forms edges using $K$ nearest neighbor in the feature space (Section 3.1). All the resulting graphs are then processed by pipeline described in section 3.2 and 3.3.

\textbf{BEV Projection and Output:} After processing, each graph query is projected back onto the BEV grid. For each query set, the outputs are aggregated via mean pooling to produce a BEV feature map.  Maps from all sets are concatenated along the channel dimension to form the final GQN output, representing a multi-set granularity fusion of relational and contextual information.
 
\textbf{Query Supervision:} Each graph query includes a global vector encoding its object-level hypothesis. These vectors are reused by the downstream detection head for classification and bounding-box regression. While the detection head is described in next section, this reuse provides a direct learning signal for the queries, ensuring that the learned graph features are grounded in object-level predictions.

GQN is designed to be modular. Although this paper uses a three-set query pyramid as a representative configuration, the framework can generalize to alternative designs or fixed configurations. A sensitivity analysis is included in the ablation study.

\section{Unified Reasoning Architecture (URA)}

Building on the modular design of GQN, the Unified Reasoning architecture (URA) integrates it into a full radar perception pipeline. Radar data is inherently sparse, noisy, and ambiguous, making it challenging for standalone modules to support robust detection performance. While graph features capture local structure and interactions between objects, but they may lack global awareness. Conversely, global modules may overlook fine-grained and inter-object interactions. Additionally, temporal fusion is essential to reduce frame-level noise and support object completeness. URA addresses these challenges by combining temporal fusion, graph processing via GQN, and global modeling into a unified architecture.

As shown in Figure \ref{full}, radar point-cloud input is first encoded into BEV feature maps by a Pillar FeatureNet backbone (section 3.1). These maps are passed into a radar-based Temporal fusion modules (e.g., ConvGRU \cite{decourt2024recurrent}, AttentiveGRU \cite{saini2025attentivegru}), to aggregate information across frames to suppress noise and improve signal stability. This early denoising step prevents graph queries from attending to clutter and allows global module to focus on spatial context rather than temporal alignment. 

The architecture then branches into two parallel reasoning pathways to ensure that local graph reasoning and global reasoning operate not only as independent and complementary processes, but are also in alignment:

\begin{itemize}
\item \textbf{Graph Reasoning Pathway}: Temporally fused features are processed by GQN pipeline (section 3). A skip connection combines the GQN input with its output via MLP (MLP1), enabling spatial interpretation by anchoring relative dependencies to the original fused features.

\item \textbf{Global Reasoning Pathway}: In parallel, temporally fused BEV features are processed by a global feature extractor, such as a Deformable DETR encoder \cite{zhu2020deformable} or FPN encoder \cite{lang2019pointpillars}, followed by a Centerpoint Decoder \cite{Saini_2024_CVPR}. In our design, the learned Deformable-DETR style decoder queries of this module are replaced with the global vectors of graph queries, anchoring global reasoning to the same object hypotheses discovered by GQN.

\end{itemize}

 Because both pathways use shared query vectors, URA promotes representational coherence. Their outputs are merged using soft attention, where a lightweight MLP (MLP2) with softmax activation predicts per-pixel weights for adaptive fusion. This mechanism balances relational cues from GQN with global scene context, rather than concatenating features.

Finally, the fused BEV features are passed to a CenterPoint detection head \cite{tian2019fcos}, following trends in prior radar-based works \cite{Saini_2024_CVPR,nabati2021centerfusion,cheng2024centerradarnet}, which performs object classification and bounding box regression using spatial center heatmaps and offers strong robustness to sparse and noisy signals. 

The detection head operates atop the fused outputs from the Centerpoint decoder and GQN. By design, the CenterPoint decoder \cite{Saini_2024_CVPR} fuses decoder queries with spatial centers of objects, allowing query learning signals to propagate directly from the detection head. Because, in our design, decoder queries are reused from the global vectors of graph queries, their fusion with center features provides object-centric supervision during graph construction and enforces semantic alignment between graph reasoning and detection outputs, mitigating the risk of latching onto noise artifacts.

In summary, URA unifies graph, global, and temporal feature extraction around a shared object-centric foundation. Its modularity is demonstrated by interchangeable pathways: Temporal Fusion (ConvGRU, AttentiveGRU, etc.), Global Feature Extractor (FPN, Transformers, etc.), and the optional GQN pathway, that can be swapped without altering the detection head or loss, enabling consistent integration across diverse baselines.

\begin{table*}[t]
\centering
\caption{Benchmark on the nuScenes dataset \cite{caesar2020nuscenes}. Annotations Rel.- Relative, Tf.- Transformer}
\setlength{\tabcolsep}{4.5pt} 
\begin{tabular}{lcccccccccc}
\toprule
\textbf{Object Detector} & \textbf{mAP (\%)} $\uparrow$ & \textbf{NDS} $\uparrow$ & \textbf{mATE}  & \textbf{mAOE}  & \textbf{mAVE}   &  \textbf{mASE} & \textbf{FPS} & \textbf{rel. mAP} $\uparrow$ & \textbf{+ GFLOPs} \\
\midrule
Radar-PointGNN \cite{svenningsson2021radar}  & 13.7 & -- & -- & -- & -- & --  & --& -- & -- \\

KPConvPillars \cite{ulrich2022improved}      & 24.4 & -- & -- & -- & -- & --  & --& -- & -- \\
KPPillarsBEV \cite{kohler2023improved}      & 26.4 & -- & -- & -- & -- & --  & --& -- & -- \\
\midrule
PointPillars \cite{lang2019pointpillars}     & 22.0 & 26.9 & 0.766 & 0.317 & 1.693  & 0.169& 42 & - & 40.6 \\
\hspace{0.8em}+ GQN                           & \textbf{33.7} & \textbf{34.6} & 0.510 & 0.266 & 1.572 & 0.490 & 29 & \textbf{+53.1\%} & +17.3 \\
\midrule
CenterPoint Tf. \cite{Saini_2024_CVPR}       & 32.4 & 32 & 0.532 & 0.053 & 1.562  & 0.467& 35 & - & 61.3 \\
\hspace{0.8em}+ GQN                           & \textbf{38.4} & \textbf{35.1} & 0.504 & 0.055 & 1.590 & 0.487 & 23 & \textbf{+18.5\%} & +17.3 \\
\midrule
AttentiveGRU \cite{saini2025attentivegru}    & 39.2 & 34.8 & 0.517 & 0.056 & 1.720  & 0.473 & 28& - & 51.3 \\
\hspace{0.8em}+ GQN                           & \textbf{42.4} & \textbf{37.2} & 0.500 & 0.055 & 1.585  & 0.513  & 21 & \textbf{+8.2\%} & +17.3 \\
\bottomrule
\end{tabular}
\label{nu}
\end{table*}

\section{Experiments}

\subsection{Datasets}

Given the large domain gap in terms of sensor characteristics between different radar sensors and levels of radar data, we tested our model on two different datasets. The nuScenes dataset restricts the number of submissions for its test dataset, prompting us to perform our comprehensive comparison with all relevant works, reimplemented from the literature, on its official validation dataset, following the test approach in \cite{ulrich2022improved}. Furthermore, although our network is capable of detecting object types such as bikes and pedestrians, it is important to note that the current state-of-the-art for radar-only detectors on the nuScenes dataset predominantly focuses on evaluating models based on their performance in detecting the 'car' class. This has become a de facto standard evaluation approach, as reflected in the works \cite{ulrich2022improved,kohler2023improved,svenningsson2021radar,meyer2021graph,Saini_2024_CVPR,nobis2021kernel}. Consequently, in line with this common practice in literature, our quantitative evaluation on the nuScenes dataset focuses primarily on the "car" class, which accounts for the majority of the bounding boxes in the dataset. In our work, we use standard validation and training split of public nuScene dataset.

To nevertheless analyze how graph reasoning interacts with different object categories, we conduct additional studies on a proprietary automotive-grade dataset (permitted under CVF guidelines) introduced in \cite{braun2021quantification}. This dataset features broader class diversity through explicit annotation of object motion (moving vs.\ stationary) and physical scale (large vs.\ small), including vulnerable road users. We use 21,766 scenes for training and 9,294 for testing.

\subsection{Implementation Details}
\textbf{Training Setup:}
All models are trained using the Adam optimizer with a learning rate of $1 \times 10\textsuperscript{-4}$ for 29 epochs, followed by a final fine-tuning epoch at $1 \times 10\textsuperscript{-5}$. Training is performed on a single Nvidia 3090 GPU with a batch size of $1$. To address class imbalance, we use a sigmoid focal-based loss \cite{lin2017focal} for classification and a smooth L1 loss for bounding box regression. All feature dimensions are fixed to $d=64$ across the backbone, temporal module, GQN, and global module, enabling seamless composition without requiring additional projection layers and balancing model capacity and efficiency. 

\textbf{URA Modules:}
For URA (section 5), all temporal modules (e.g., ConvGRU, AttentiveGRU), global modules (FPN, transformers), and centerpoint decoder and head adopt default configurations from their respective cited works. Sinusoidal positional embeddings follow the standard formulation \cite{vaswani2017attention} with a frequency base of 100 to better capture spatial variations in radar data.

\textbf{GQN Configuration:}
GQN employs three sets of learnable graph queries, each containing $M=32$   queries (total $\tau = 96$). For multi-set design, queries from each set samples different ratios of BEV feature locations:  10\%, 20\%, 30\% respectively. Within each set, a graph is constructed for each query over its sampled nodes using KNN in the state feature space, with $ K = 4, 8, 12$ scaled proportionally to its set-specific sampling density. This multi-set design, along with its parameter choices, reflects a trade-off between performance and FLOPs, and is empirically motivated via sensitivity analysis in Section 5.4.

\textbf{MLP Dimensions:}
The EdgeFocus operator uses MLPs $\phi(.)$ and $\rho(.)$ with configuration  64*2$\rightarrow$64$\rightarrow$64. The DeepContext pooling module uses $L=6$ self-attention layers and $\eta(.)$ configured as: 64*2$\rightarrow$64$\rightarrow$64. For the attention-weighted fusion module, MLP1 follows 64*5$\rightarrow$128$\rightarrow$64, and MLP2 follows 64*2$\rightarrow$64$\rightarrow$2, followed by a Softmax activation to compute fusion weights.

\subsection{Results}

We evaluate the effectiveness of GQN on the nuScenes dataset by integrating it into three representative radar detection baselines: PointPillars \cite{lang2019pointpillars}, CenterPoint Transformer \cite{Saini_2024_CVPR} and  Attentive GRU \cite{saini2025attentivegru}, all implemented using URA introduced in Section 5. These works span a spectrum of architectural diversity, from minimal FPNs to advanced temporal-transformer hybrids. To contextualize the improvements, we compare these GQN-enhanced models to prior radar detectors reported on nuScenes, including:

\textbf{(1)} \textit{Graph-based models:} Radar-PointGNN \cite{svenningsson2021radar}, KPConvPillars \cite{ulrich2022improved}, KPPillarsBEV  \cite{kohler2023improved};

\textbf{(2)} \textit{Vanilla versions} of aforementioned PointPillars, CenterPoint Transformer, and AttentiveGRU. They serve as comparison points and integration candidates for GQN.

Table \ref{nu} reports the full set of nuScenes metrics \cite{caesar2020nuscenes} (mAP, mATE, etc.), along with FPS (3090 GPU) and FLOPs.
 GQN-enhanced variants consistently outperform both their corresponding vanilla versions and prior radar detectors.

 The largest gain is observed when GQN is added to PointPillars (+53.1\%), which lacks temporal fusion and global context modules, highlighting the standalone strength of GQN. Notably, KPConvPillars also extends the PointPillars baseline, yet our GQN-enhanced variant achieves higher performance with only a lightweight FPN. Even when integrated into more advanced models like CenterPoint Transformer (+18.5\%) and AttentiveGRU (+8.2\%), both featuring temporal fusion and global attention, GQN provides further improvements, underscoring its complementarity role to existing reasoning modules. In terms of FLOPs, GQN adds a fixed overhead of 17.3 GFLOPs across all URA baselines as shown in Table \ref{nu}. This modest increase reflects the cost of graph reasoning, but at $\sim$21 FPS, above the nuScenes radar rate (13 Hz, 2 Hz keyframes), the framework supports real-time deployment.

 Importantly, GQN avoids full-graph construction by using 96 query-specific graphs over 10-30\% of BEV scene with $K=4,8,12$ avoiding full-scene graphs with $K=20$. This design leads to over 80\% lower peak compute during graph processing ($ \mathcal{O}(NK)$). Graph construction ($ \mathcal{O}(NlogN + NK)$) is similarly efficient, with over 80\% lower peak cost. These savings make GQN both scalable and a modular component, enabling real-time deployment without compromising performance.

Overall, the results demonstrate that GQN enhances diverse detector architectures, improves performance beyond strong transformer and full-graph baselines, and maintain efficiency through selective, query-centric graph reasoning.

\subsection{Ablation Studies}

\begin{figure}[]
    \def\svgwidth{\linewidth}
    \centering
    \fontsize{6}{8}\selectfont
    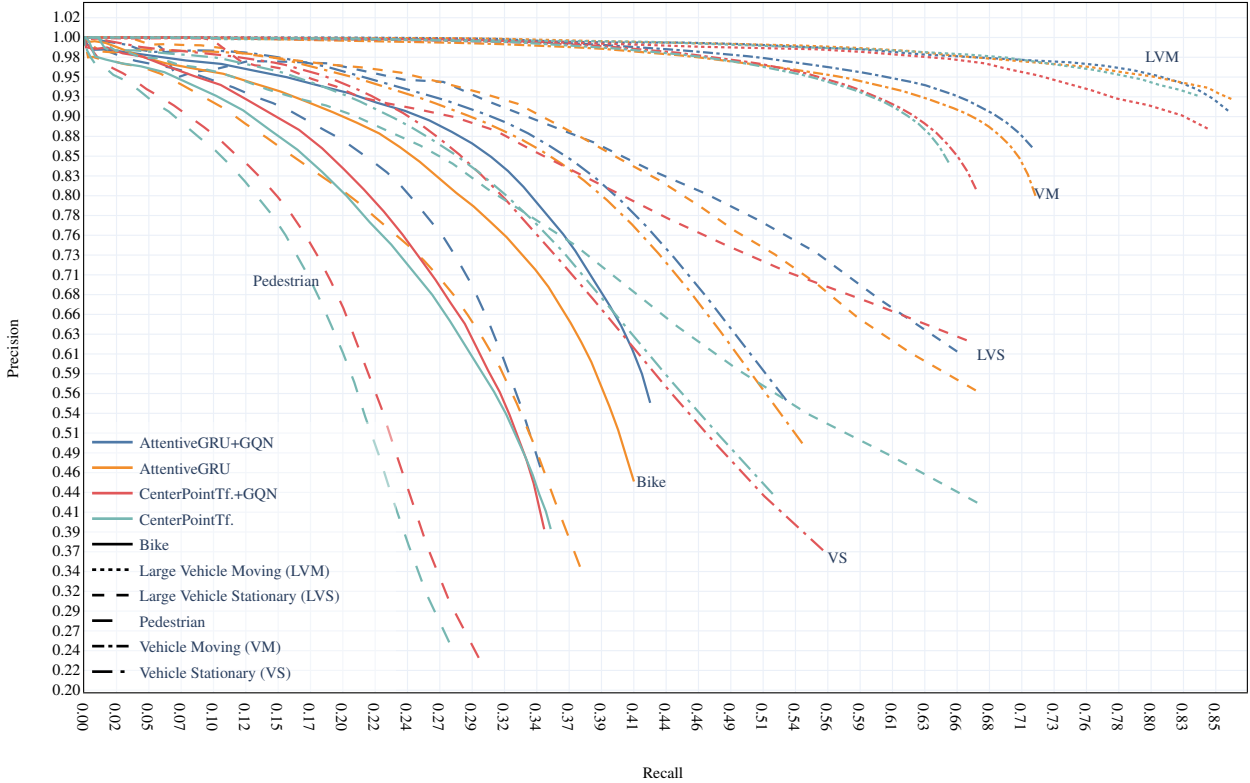
    \caption{Precision-Recall curves for secondary dataset \cite{braun2021quantification}}
    \label{pr}
\end{figure}

\textbf{Class Sensitivity:} To assess how graph reasoning interacts with diverse class categories, we compare baselines with and without GQN on an additional dataset \cite{braun2021quantification}, reporting the class-wise AUC from precision-recall curves. As shown in Figure~\ref{pr}, GQN enhanced variants yields the largest gains for safety-critical classes such as pedestrians and bicycles, as well as for stationary vehicles, which often exhibit sparse radar returns or weak Doppler signatures where motion-based detection is unreliable. GQN amplifies structural cues in these cases, while for moving vehicles, where Doppler already provides strong supervision, it offers smaller but consistent gains. These results indicate that the FLOP overhead is most valuable in the most challenging and safety-critical setting: vulnerable road user detection.

\textbf{Component-wise ablation:} Table~\ref{tab:ablation} isolates the impact of key GQN components using AttentiveGRU as the representative backbone. Replacing the EdgeFocus with simple attention-based summation over node features (no edge features) reduces AP\textsubscript{4.0} by 1.8\%. Removing DeepContext, by replacing its output with a zero tensor, also leads to a drop, and jointly disabling both yields the largest performance drop (-4.1\%). Interestingly, removing entire GQN performs better than using a partially active version with broken message passing, suggesting that individual modules contribute synergistically to performance.

\begin{table}[h]
\centering
\caption{Component-wise ablation (nuScenes), Full: AttentiveGRU+GQN (EdgeFocus+DeepContext). 
-Edge: replace EdgeFocus, -Ctx: replace DeepContext, -Both: replace both. 
Base: AttentiveGRU only }
\label{tab:ablation}
\begin{tabular}{lccccc}
\toprule
\textbf{Variant} & Full & -Edge & -Ctx & -Both & Base \\
\midrule
AP\textsubscript{4.0} (\%) & \textbf{53.9} & 52.1 & 51.8 & 49.8 & 50.4 \\
\bottomrule
\end{tabular}
\end{table}

\textbf{K Value Sensitivity:} To isolate the effect of $K$, we fix the sampling ratio to 20\% ($\tau$=96) and vary $K$ independently. As shown in Table \ref{tab:k_ablation}, increasing $K$  improves performance, peaking at $K$=16, suggesting that a higher neighborhood size is beneficial when fewer nodes are sampled. However, further increase to $K$=20 degrades performance and increases FLOPs, suggesting over-connected graphs at a given sampling may saturate structural information. This motivates scaling $K$ proportionally to sampling ratio.

\begin{table}[h]
\centering
\caption{K-value sensitivity at fixed 20\% sampling (nuScenes)}
\label{tab:k_ablation}
\small
\begin{tabular}{lcccccc}
\toprule
\textbf{K Value} & 2 & 4 & 8 & 12 & 16 & 20 \\
\midrule
AP\textsubscript{4.0} (\%) & 51.5 & 51.54 & 52.9 & 52.7 & \textbf{53.3} & 51.46 \\
GFLOPs & 61.5 & 62.7 & 65.0 & 67.4 & 69.8 & 72.1 \\
\bottomrule
\end{tabular}
\end{table}

\textbf{Sampling Density:} Table \ref{tab:sampling_ablation} reports the performance and cost of the single and two-set variants ($\tau$=96 for all) with increasing BEV sampling ratios on nuScenes. Following $K$ sensitivity analysis, $K$ is scaled with sampling ratios to maintain sufficient neighborhood size while balancing FLOPs (e.g., $K$=8 for 20\%). AP peaks at 20\% sampling, and beyond 30\%, performance saturates while FLOPs rise sharply, suggesting a practical upper bound. Two-set variants (e.g., 10,20\%) do not outperform single-scale setups, indicating that two granularities sets may be insufficient. These trends motivate our final three-set design.

\begin{table}[h]
\centering
\caption{Sampling ratio (S) with scaled $K$ values (nuScenes)}
\label{tab:sampling_ablation}
\setlength{\tabcolsep}{3pt} 
\renewcommand{\arraystretch}{1.05}
\small
\begin{tabular}{@{}lccccc|cc@{}}
\toprule
\textbf{S} & 10\% & 20\% & 30\% & 40\% & 50\% & 10,20\% & 20,30\% \\
\textbf{K} & 4    & 8    & 12   & 16   & 20   & 4,8     & 8,12 \\
\midrule
AP\textsubscript{4.0} (\%) & 51.9 & \textbf{52.9} & 51.6 & 51.7 & 51.5 & 50.9 & 52.1 \\
GFLOPs & 57.7 & 65.0 & 76.0 & 93.3 & 114.0 & 60.8 & 70.2 \\
\bottomrule
\end{tabular}
\end{table}

These studies highlight the need to jointly tune sampling density and connectivity. The final three set design in GQN (Section 5.2) spans coarse-to-dense sampling levels with scaled $K$, enabling diverse neighborhood coverage across sparsity granularities. Unlike 
 naive multi-set variants, it avoids redundant computation and provide best tradeoff: \textbf{53.9} $AP_{4.0}$ at \textbf{68.6} GFLOPs. While a full grid search across all (sampling,$K$,sets) combinations is intractable, our sensitivity analyses provide the rationale behind this design.

 \section{Conclusion}

In this paper, we introduced the Graph Query Network (GQN), a novel framework that leverages dynamic graph queries to extract relational and contextual features for radar-based object detection. GQN instantiates graph queries over BEV features through attention-guided sampling, followed by relational reasoning (EdgeFocus), inter-object context reasoning (DeepContext Pooling) and coarse-to-fine representation though multi-set design. By localizing graph reasoning through sampled queries, GQN modularizes graph processing into a plug-and-play component compatible with diverse backbones. We integrated GQN into a Unified Reasoning Architecture (URA) combining temporal, graph, and global modules. Experiments on two radar datasets, including sampling and K-sensitivity studies and ablations, validate its effectiveness, robustness, and efficiency. Our findings highlight the potential of structured, query-centric graph reasoning to advance radar perception.

\bibliographystyle{unsrt}  
\bibliography{references}  

@inproceedings{saini2025attentivegru,
  title={AttentiveGRU: Recurrent Spatio-Temporal Modeling for Advanced Radar-Based BEV Object Detection},
  author={Saini, Loveneet and Meuter, Mirko and Tercan, Hasan and Meisen, Tobias},
  booktitle={Proceedings of the Computer Vision and Pattern Recognition Conference},
  pages={2415--2424},
  year={2025}
}

@inproceedings{lin2017focal,
  title={Focal loss for dense object detection},
  author={Lin, Tsung-Yi and Goyal, Priya and Girshick, Ross and He, Kaiming and Doll{\'a}r, Piotr},
  booktitle={Proceedings of the IEEE international conference on computer vision},
  pages={2980--2988},
  year={2017}
}

@inproceedings{tian2019fcos,
  title={Fcos: Fully convolutional one-stage object detection},
  author={Tian, Zhi and Shen, Chunhua and Chen, Hao and He, Tong},
  booktitle={Proceedings of the IEEE/CVF international conference on computer vision},
  pages={9627--9636},
  year={2019}
}

@inproceedings{carion2020end,
  title={End-to-end object detection with transformers},
  author={Carion, Nicolas and Massa, Francisco and Synnaeve, Gabriel and Usunier, Nicolas and Kirillov, Alexander and Zagoruyko, Sergey},
  booktitle={European conference on computer vision},
  pages={213--229},
  year={2020},
  organization={Springer}
}

@article{zhu2020deformable,
  title={Deformable detr: Deformable transformers for end-to-end object detection},
  author={Zhu, Xizhou and Su, Weijie and Lu, Lewei and Li, Bin and Wang, Xiaogang and Dai, Jifeng},
  journal={arXiv preprint arXiv:2010.04159},
  year={2020}
}

@inproceedings{svenningsson2021radar,
  title={Radar-pointgnn: Graph based object recognition for unstructured radar point-cloud data},
  author={Svenningsson, Peter and Fioranelli, Francesco and Yarovoy, Alexander},
  booktitle={2021 IEEE Radar Conference (RadarConf21)},
  pages={1--6},
  year={2021},
  organization={IEEE}
}

@inproceedings{saini2023graph,
  title={Graph Neural Networks for Object Type Classification Based on Automotive Radar Point Clouds and Spectra},
  author={Saini, Loveneet and Acosta, Axel and Hakobyan, Gor},
  booktitle={ICASSP 2023-2023 IEEE International Conference on Acoustics, Speech and Signal Processing (ICASSP)},
  pages={1--5},
  year={2023},
  organization={IEEE}
}

@inproceedings{braun2021quantification,
  title={Quantification of uncertainties in deep learning-based environment perception},
  author={Braun, Marco and Luszek, Moritz and Siegemund, Jan and Kollek, Kevin and Kummert, Anton},
  booktitle={2021 IEEE International Conference on Omni-Layer Intelligent Systems (COINS)},
  pages={1--8},
  year={2021},
  organization={IEEE}
}

@inproceedings{lin2017feature,
  title={Feature pyramid networks for object detection},
  author={Lin, Tsung-Yi and Doll{\'a}r, Piotr and Girshick, Ross and He, Kaiming and Hariharan, Bharath and Belongie, Serge},
  booktitle={Proceedings of the IEEE conference on computer vision and pattern recognition},
  pages={2117--2125},
  year={2017}
}

@misc{li2022bevformer,
      title={BEVFormer: Learning Bird's-Eye-View Representation from Multi-Camera Images via Spatiotemporal Transformers}, 
      author={Zhiqi Li and Wenhai Wang and Hongyang Li and Enze Xie and Chonghao Sima and Tong Lu and Qiao Yu and Jifeng Dai},
      year={2022},
      eprint={2203.17270},
      archivePrefix={arXiv},
      primaryClass={cs.CV}
}

@inproceedings{K_hler_2023,
   title={Improved Multi-Scale Grid Rendering of Point Clouds for Radar Object Detection Networks},
   url={http://dx.doi.org/10.23919/FUSION52260.2023.10224223},
   DOI={10.23919/fusion52260.2023.10224223},
   booktitle={2023 26th International Conference on Information Fusion (FUSION)},
   publisher={IEEE},
   author={Köhler, Daniel and Quach, Maurice and Ulrich, Michael and Meinl, Frank and Bischoff, Bastian and Blume, Holger},
   year={2023},
   month=jun }

@article{tan20223d,
  title={3d object detection for multi-frame 4d automotive millimeter-wave radar point cloud},
  author={Tan, Bin and Ma, Zhixiong and Zhu, Xichan and Li, Sen and Zheng, Lianqing and Chen, Sihan and Huang, Libo and Bai, Jie},
  journal={IEEE Sensors Journal},
  year={2022},
  publisher={IEEE}
}

@misc{ulrich2022improved,
      title={Improved Orientation Estimation and Detection with Hybrid Object Detection Networks for Automotive Radar}, 
      author={Michael Ulrich and Sascha Braun and Daniel Köhler and Daniel Niederlöhner and Florian Faion and Claudius Gläser and Holger Blume},
      year={2022},
      eprint={2205.02111},
      archivePrefix={arXiv},
      primaryClass={cs.CV}
}

@misc{caesar2020nuscenes,
      title={nuScenes: A multimodal dataset for autonomous driving}, 
      author={Holger Caesar and Varun Bankiti and Alex H. Lang and Sourabh Vora and Venice Erin Liong and Qiang Xu and Anush Krishnan and Yu Pan and Giancarlo Baldan and Oscar Beijbom},
      year={2020},
      eprint={1903.11027},
      archivePrefix={arXiv},
      primaryClass={cs.LG}
}

@inproceedings{lang2019pointpillars,
  title={Pointpillars: Fast encoders for object detection from point clouds},
  author={Lang, Alex H and Vora, Sourabh and Caesar, Holger and Zhou, Lubing and Yang, Jiong and Beijbom, Oscar},
  booktitle={Proceedings of the IEEE/CVF conference on computer vision and pattern recognition},
  pages={12697--12705},
  year={2019}
}

@article{vaswani2017attention,
  title={Attention is all you need},
  author={Vaswani, Ashish and Shazeer, Noam and Parmar, Niki and Uszkoreit, Jakob and Jones, Llion and Gomez, Aidan N and Kaiser, {\L}ukasz and Polosukhin, Illia},
  journal={Advances in neural information processing systems},
  volume={30},
  year={2017}
}

@inproceedings{li2022exploiting,
  title={Exploiting temporal relations on radar perception for autonomous driving},
  author={Li, Peizhao and Wang, Pu and Berntorp, Karl and Liu, Hongfu},
  booktitle={Proceedings of the IEEE/CVF Conference on Computer Vision and Pattern Recognition},
  pages={17071--17080},
  year={2022}
}

@inproceedings{yataka2024radar,
  title={Radar Perception with Scalable Connective Temporal Relations for Autonomous Driving},
  author={Yataka, Ryoma and Wang, Pu and Boufounos, Petros and Takahashi, Ryuhei},
  booktitle={ICASSP 2024-2024 IEEE International Conference on Acoustics, Speech and Signal Processing (ICASSP)},
  pages={13266--13270},
  year={2024},
  organization={IEEE}
}

@inproceedings{meyer2021graph,
  title={Graph convolutional networks for 3d object detection on radar data. In 2021 IEEE},
  author={Meyer, Michael and Kuschk, Georg and Tomforde, Sven},
  booktitle={CVF International Conference on Computer Vision Workshops (ICCVW)},
  pages={3053--3062},
  year={2021}
}

@InProceedings{Saini_2024_CVPR,
    author    = {Saini, Loveneet and Su, Yu and Tercan, Hasan and Meisen, Tobias},
    title     = {CenterPoint Transformer for BEV Object Detection with Automotive Radar},
    booktitle = {Proceedings of the IEEE/CVF Conference on Computer Vision and Pattern Recognition (CVPR) Workshops},
    month     = {June},
    year      = {2024},
    pages     = {4451-4460}
}

@article{lidar_survey_2023,
  title={A Survey on LiDAR Sensing and Object Detection for Autonomous Driving},
  author={Li, X. and Sun, Y. and Wang, J.},
  journal={Sensors},
  volume={23},
  number={6},
  pages={3223},
  year={2023},
  publisher={MDPI},
  url={https://www.mdpi.com/1424-8220/23/6/3223}
}

@article{radar_classification_survey_2020,
  title={A Survey of Radar-Based Automotive Classification Methods},
  author={Major, B. and Fontijne, D. and Ansari, A. and others},
  journal={arXiv preprint arXiv:2006.05485},
  year={2020},
  url={https://arxiv.org/abs/2006.05485}
}

@misc{4d_radar_market_report_2024,
  title={Automotive 4D Imaging Radar Market Report 2024-2031},
  author={Proficient Market Insights},
  howpublished={\url{https://www.proficientmarketinsights.com/market-reports/automotive-4d-imaging-radar-market-1534}},
  year={2024}
}

@article{patole2017automotive,
  title={Automotive radar: A review of signal processing techniques},
  author={Patole, S. M. and Torlak, M. and Wang, D. and Ali, M.},
  journal={IEEE Signal Processing Magazine},
  volume={34},
  number={2},
  pages={22--35},
  year={2017},
  publisher={IEEE}
}

@article{major2020survey,
  title={Radar Object Detection: A Survey of Classical and Deep Learning Approaches},
  author={Major, B. and Fontijne, D. and Ansari, A. and others},
  journal={arXiv preprint arXiv:2006.05485},
  year={2020}
}

@article{3d_radar_remote_sensing_2019,
  title={Automotive Radar: From First Efforts to Future Systems},
  author={Hasch, J. and Topak, E. and Schnabel, R. and Zwick, T. and Weigel, R. and Waldschmidt, C.},
  journal={Remote Sensing},
  volume={11},
  number={10},
  pages={1156},
  year={2019},
  publisher={MDPI},
  url={https://www.mdpi.com/2072-4292/11/10/1156}
}

@article{bai2021radar,
  title={Radar Transformer: An Object Classification Network Based on 4D MMW Imaging Radar},
  author={Bai, Jie and Zheng, Lianqing and Li, Sen and Tan, Bin and Chen, Sihan and Huang, Libo},
  journal={Sensors},
  volume={21},
  number={11},
  pages={3854},
  year={2021},
  publisher={MDPI},
  doi={10.3390/s21113854}
}

@inproceedings{yataka2024sira,
  title={SIRA: Scalable Inter-frame Relation and Association for Radar Perception},
  author={Yataka, Ryoma and Wang, Pu and Boufounos, Petros and Takahashi, Ryuhei},
  booktitle={Proceedings of the IEEE/CVF Conference on Computer Vision and Pattern Recognition},
  pages={15024--15034},
  year={2024}
}

@inproceedings{fent2023radargnn,
  title={RadarGNN: Transformation invariant graph neural network for radar-based perception},
  author={Fent, Felix and Bauerschmidt, Philipp and Lienkamp, Markus},
  booktitle={Proceedings of the IEEE/CVF Conference on Computer Vision and Pattern Recognition},
  pages={182--191},
  year={2023}
}

@inproceedings{wu2024point,
  title={Point transformer v3: Simpler faster stronger},
  author={Wu, Xiaoyang and Jiang, Li and Wang, Peng-Shuai and Liu, Zhijian and Liu, Xihui and Qiao, Yu and Ouyang, Wanli and He, Tong and Zhao, Hengshuang},
  booktitle={Proceedings of the IEEE/CVF Conference on Computer Vision and Pattern Recognition},
  pages={4840--4851},
  year={2024}
}

@article{battaglia2018relational,
  title={Relational inductive biases, deep learning, and graph networks},
  author={Battaglia, Peter W and Hamrick, Jessica B and Bapst, Victor and Sanchez-Gonzalez, Alvaro and Zambaldi, Vinicius and Malinowski, Mateusz and Tacchetti, Andrea and Raposo, David and Santoro, Adam and Faulkner, Ryan and others},
  journal={arXiv preprint arXiv:1806.01261},
  year={2018}
}

@article{wang2019dynamic,
  title={Dynamic graph cnn for learning on point clouds},
  author={Wang, Yue and Sun, Yongbin and Liu, Ziwei and Sarma, Sanjay E and Bronstein, Michael M and Solomon, Justin M},
  journal={ACM Transactions on Graphics (tog)},
  volume={38},
  number={5},
  pages={1--12},
  year={2019},
  publisher={Acm New York, NY, USA}
}

@article{decourt2024recurrent,
  title={A recurrent CNN for online object detection on raw radar frames},
  author={Decourt, Colin and VanRullen, Rufin and Salle, Didier and Oberlin, Thomas},
  journal={IEEE Transactions on Intelligent Transportation Systems},
  year={2024},
  publisher={IEEE}
}

@article{musiat2024radarpillars,
  title={RadarPillars: Efficient Object Detection from 4D Radar Point Clouds},
  author={Musiat, Alexander and Reichardt, Laurenz and Schulze, Michael and Wasenmüller, Oliver},
  journal={arXiv preprint arXiv:2408.05020},
  year={2024},
  url={https://arxiv.org/abs/2408.05020}
}

@article{lee2020deep,
  title={Deep Learning on Radar Centric 3D Object Detection},
  author={Lee, Seungjun},
  journal={arXiv preprint arXiv:2003.00851},
  year={2020},
  url={https://arxiv.org/abs/2003.00851}
}

@inproceedings{peng2024transloc4d,
  title={TransLoc4D: Transformer-based 4D Radar Place Recognition},
  author={Peng, Guohao and Li, Heshan and Zhao, Yangyang and Zhang, Jun and Wu, Zhenyu and Zheng, Pengyu and Wang, Danwei},
  booktitle={Proceedings of the IEEE/CVF Conference on Computer Vision and Pattern Recognition (CVPR)},
  pages={17595--17605},
  year={2024}
}

@article{jiang2023mlrt,
  title={Multi-Task Learning Radar Transformer (MLRT): A Personal Identification and Fall Detection Network Based on IR-UWB Radar},
  author={Jiang, Xikang and Zhang, Lin and Li, Lei},
  journal={Sensors},
  volume={23},
  number={12},
  pages={5632},
  year={2023},
  publisher={MDPI},
  doi={10.3390/s23125632}
}

@inproceedings{shi2020point,
  title={Point-GNN: Graph Neural Network for 3D Object Detection in a Point Cloud},
  author={Shi, Weijing and Rajkumar, Ragunathan (Raj)},
  booktitle={Proceedings of the IEEE/CVF Conference on Computer Vision and Pattern Recognition (CVPR)},
  pages={1711--1719},
  year={2020},
  doi={10.1109/CVPR42600.2020.00179}
}

@inproceedings{song2024graphbev,
  title={GraphBEV: Towards Robust BEV Feature Alignment for Multi-Modal 3D Object Detection},
  author={Song, Ziying and Yang, Lei and Xu, Shaoqing and Liu, Lin and Xu, Dongyang and Jia, Caiyan and Jia, Feiyang and Wang, Li},
  booktitle={Proceedings of the European Conference on Computer Vision (ECCV)},
  year={2024}
}

@inproceedings{nabati2021centerfusion,
  title={Centerfusion: Center-based radar and camera fusion for 3d object detection},
  author={Nabati, Ramin and Qi, Hairong},
  booktitle={Proceedings of the IEEE/CVF winter conference on applications of computer vision},
  pages={1527--1536},
  year={2021}
}

@inproceedings{cheng2024centerradarnet,
  title={Centerradarnet: Joint 3d object detection and tracking framework using 4d fmcw radar},
  author={Cheng, Jen-Hao and Kuan, Sheng-Yao and Liu, Hou-I and Latapie, Hugo and Liu, Gaowen and Hwang, Jenq-Neng},
  booktitle={2024 IEEE International Conference on Image Processing (ICIP)},
  pages={998--1004},
  year={2024},
  organization={IEEE}
}

@article{han2022vision,
  title={Vision gnn: An image is worth graph of nodes},
  author={Han, Kai and Wang, Yunhe and Guo, Jianyuan and Tang, Yehui and Wu, Enhua},
  journal={Advances in neural information processing systems},
  volume={35},
  pages={8291--8303},
  year={2022}
}

@article{hasch2012millimeter,
  title={Millimeter-wave technology for automotive radar sensors in the 77 GHz frequency band},
  author={Hasch, J{\"u}rgen and Topak, Eray and Schnabel, Raik and Zwick, Thomas and Weigel, Robert and Waldschmidt, Christian},
  journal={IEEE transactions on microwave theory and techniques},
  volume={60},
  number={3},
  pages={845--860},
  year={2012},
  publisher={IEEE}
}

@inproceedings{kohler2023improved,
  title={Improved multi-scale grid rendering of point clouds for radar object detection networks},
  author={K{\"o}hler, Daniel and Quach, Maurice and Ulrich, Michael and Meinl, Frank and Bischoff, Bastian and Blume, Holger},
  booktitle={2023 26th International Conference on Information Fusion (FUSION)},
  pages={1--8},
  year={2023},
  organization={IEEE}
}

@article{nobis2021kernel,
  title={Kernel point convolution LSTM networks for radar point cloud segmentation},
  author={Nobis, Felix and Fent, Felix and Betz, Johannes and Lienkamp, Markus},
  journal={Applied Sciences},
  volume={11},
  number={6},
  pages={2599},
  year={2021},
  publisher={MDPI}
}

@article{velivckovic2017graph,
  title={Graph attention networks},
  author={Veli{\v{c}}kovi{\'c}, Petar and Cucurull, Guillem and Casanova, Arantxa and Romero, Adriana and Lio, Pietro and Bengio, Yoshua},
  journal={arXiv preprint arXiv:1710.10903},
  year={2017}
}

@article{wang2023sat,
  title={SAT-GCN: Self-attention graph convolutional network-based 3D object detection for autonomous driving},
  author={Wang, Li and Song, Ziying and Zhang, Xinyu and Wang, Chenfei and Zhang, Guoxin and Zhu, Lei and Li, Jun and Liu, Huaping},
  journal={Knowledge-Based Systems},
  volume={259},
  pages={110080},
  year={2023},
  publisher={Elsevier}
}

@article{song2023vp,
  title={VP-Net: Voxels as points for 3-D object detection},
  author={Song, Ziying and Wei, Haiyue and Jia, Caiyan and Xia, Yongchao and Li, Xiaokun and Zhang, Chao},
  journal={IEEE Transactions on Geoscience and Remote Sensing},
  volume={61},
  pages={1--12},
  year={2023},
  publisher={IEEE}
}

@inproceedings{yang2018graph,
  title={Graph r-cnn for scene graph generation},
  author={Yang, Jianwei and Lu, Jiasen and Lee, Stefan and Batra, Dhruv and Parikh, Devi},
  booktitle={Proceedings of the European conference on computer vision (ECCV)},
  pages={670--685},
  year={2018}
}

@inproceedings{song2023graphalign,
  title={Graphalign: Enhancing accurate feature alignment by graph matching for multi-modal 3d object detection},
  author={Song, Ziying and Wei, Haiyue and Bai, Lin and Yang, Lei and Jia, Caiyan},
  booktitle={Proceedings of the IEEE/CVF international conference on computer vision},
  pages={3358--3369},
  year={2023}
}

@article{song2023graphalign++,
  title={GraphAlign++: An accurate feature alignment by graph matching for multi-modal 3D object detection},
  author={Song, Ziying and Jia, Caiyan and Yang, Lei and Wei, Haiyue and Liu, Lin},
  journal={IEEE Transactions on Circuits and Systems for Video Technology},
  volume={34},
  number={4},
  pages={2619--2632},
  year={2023},
  publisher={IEEE}
}

@article{sun2025focalad,
  title={FocalAD: Local Motion Planning for End-to-End Autonomous Driving},
  author={Sun, Bin and Zhang, Boao and Lu, Jiayi and Feng, Xinjie and Shang, Jiachen and Cao, Rui and Zheng, Mengchao and Wang, Chuanye and Yang, Shichun and Cao, Yaoguang and others},
  journal={arXiv preprint arXiv:2506.11419},
  year={2025}
}

\end{document}